\documentclass{article}

\usepackage{microtype}
\usepackage{graphicx}
\usepackage{subfigure}
\usepackage{booktabs} 
\usepackage{threeparttable}
\usepackage{hyperref}
\usepackage{multirow}
\usepackage[ruled]{algorithm2e}

\usepackage[accepted]{icml2025}

\usepackage{amsmath}
\usepackage{amssymb}
\usepackage{mathtools}
\usepackage{amsthm}

\usepackage[capitalize,noabbrev]{cleveref}

\theoremstyle{plain}

\theoremstyle{definition}

\theoremstyle{remark}

\usepackage[textsize=tiny]{todonotes}

\icmltitlerunning{CFT-RAG}

\begin{document}

\twocolumn[
\icmltitle{CFT-RAG: An Entity Tree Based Retrieval Augmented Generation Algorithm With Cuckoo Filter}



\icmlsetsymbol{equal}{*}

\begin{icmlauthorlist}
\icmlauthor{Zihang Li}{pku1}
\icmlauthor{Yangdong Ruan}{buaa}
\icmlauthor{Wenjun Liu}{pku2}
\icmlauthor{Zhengyang Wang}{pku3}
\icmlauthor{Tong Yang}{pku4}


\end{icmlauthorlist}

\icmlaffiliation{pku1}{Peking University}
\icmlaffiliation{buaa}{Beihang University}
\icmlaffiliation{pku2}{Peking University}
\icmlaffiliation{pku3}{Peking University}
\icmlaffiliation{pku4}{Peking University}

\icmlcorrespondingauthor{Tong Yang}{yangtong@pku.edu.cn}
\icmlcorrespondingauthor{Zihang Li}{lizihang@stu.pku.edu.cn}

\icmlkeywords{Retrieval-Augmented Generation, Tree-RAG, Cuckoo Filter, Knowledge Retrieval}

\vskip 0.3in
]



\printAffiliationsAndNotice{}  

\begin{abstract}
Although retrieval-augmented generation(RAG) significantly improves generation quality by retrieving external knowledge bases and integrating generated content, it faces computational efficiency bottlenecks, particularly in knowledge retrieval tasks involving hierarchical structures for Tree-RAG. This paper proposes a Tree-RAG acceleration method based on the improved Cuckoo Filter, which optimizes entity localization during the retrieval process to achieve significant performance improvements. Tree-RAG effectively organizes entities through the introduction of a hierarchical tree structure, while the Cuckoo Filter serves as an efficient data structure that supports rapid membership queries and dynamic updates. The experiment results demonstrate that our method is much faster than naive Tree-RAG while maintaining high levels of generative quality. When the number of trees is large, our method is hundreds of times faster than naive Tree-RAG. Our work is available at \href{https://github.com/TUPYP7180/CFT-RAG-2025}{https://github.com/TUPYP7180/CFT-RAG-2025}.
\end{abstract}

\section{Introduction}
\label{submission}
\begin{figure*}[h!]
    \centering
    \includegraphics[width=4.5in]{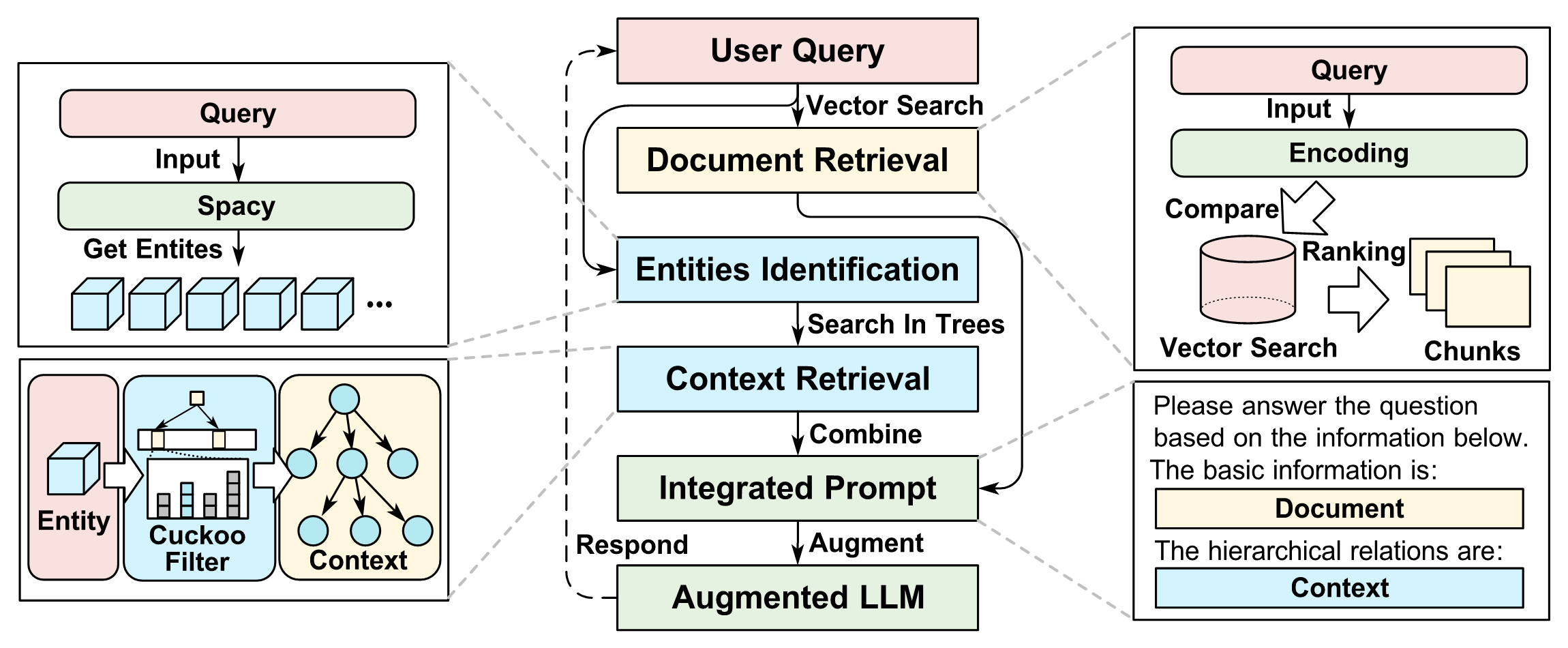}
    \caption{The workflow of CFT-RAG begins with a user query, which undergoes vector search to retrieve relevant documents. Key entities are then identified from entity trees by applying SpaCy and hierarchical tree searches. Context information related to these entities is retrieved and filtered efficiently by applying Cuckoo Filter. The retrieved context and hierarchical relationships are integrated into a comprehensive prompt, which is fed into an augmented large language model (LLM). The LLM processes this enriched prompt to generate a context-aware and accurate response to the user query.}
    \label{fig:my_label0}
\end{figure*}

In the era of information explosion, Retrieval-Augmented Generation (RAG), a technology integrating retrieval mechanisms with generative models, has gained significant attention. It allows models to draw on external knowledge bases during text generation, effectively overcoming the limitations of traditional generative models in knowledge-intensive tasks~\cite{patrick2020retrieval}. The knowledge base, a vital part of the RAG system, stores a wealth of structured and unstructured knowledge, acting as the main source of external information for the model. However, with the continuous expansion of the knowledge base and the rapid pace of knowledge update, the challenge of efficiently retrieving relevant and accurate information from it has become a major obstacle to improving the performance of RAG system. Enhancing the retrieval speed and accuracy of the knowledge base is crucial for boosting the overall performance of the RAG system. Faster and more accurate retrieval enables the model to access relevant knowledge promptly, improving response speed and the quality of generated content. In contrast, inefficient or inaccurate retrieval can result in incorrect or irrelevant outputs, degrading user experience and system usability. Therefore, exploring ways to optimize the knowledge base retrieval mechanism is of great theoretical and practical importance, and this paper will focus on this key issue.\\ \\
Knowledge bases in Retrieval-Augmented Generation (RAG) systems are mainly of three types: text-based, graph-based, and tree-based. Text-based ones store information as text, easy to manage but slow in retrieval due to complex language processing. Graph-based knowledge bases represent knowledge as graphs, excelling in handling complex relationships with relatively fast retrieval for certain queries, thanks to graph neural networks. Tree-based knowledge bases structure knowledge hierarchically. Despite text-based and graph-based knowledge bases having made good progress, the retrieval speed of all three types, especially tree-based ones, needs improvement. For RAG systems to provide faster and more accurate responses, optimizing retrieval from these knowledge bases, particularly tree-RAG, is a key research challenge.\\ \\
Tree-RAG, an extension of RAG, improves on traditional RAG frameworks by using a hierarchical tree structure to organize the retrieved knowledge, thus providing richer context and capturing complex relationships among entities. In Tree-RAG, entities are arranged hierarchically, allowing the retrieval process to more effectively traverse related entities at multiple levels. This results in enhanced response accuracy and coherence, as the tree structure maintains connections between entities that are essential for contextually rich answers~\cite{fatehkia2024t}. However, a critical limitation of Tree-RAG lies in its computational inefficiency: as the datasets and tree depth grow, the time required to locate and retrieve relevant entities within the hierarchical structure significantly increases, posing scalability challenges. This paper aims to greatly improve the retrieval efficiency of Tree-RAG without sacrificing the accuracy of the generated responses. \\ \\
To address the retrieval bottleneck, we introduce an optimized method to Tree-RAG by integrating the Cuckoo Filter, a high-performance data structure. Its basic workflow is presented in Figure \ref{fig:my_label0}. The Cuckoo Filter excels in fast membership queries and supports dynamic operations, such as element insertions and deletions, making it suitable for dynamic data management scenarios~\cite{fan2014cuckoo}. Unlike traditional filters, such as Bloom Filter, which is limited by fixed false-positive rates and lack deletion capability, the Cuckoo Filter allows for both flexible updates and reduced storage requirements. Therefore, it is particularly advantageous for handling large hierarchical datasets. The comparison experiment is designed to demonstrate that our method is significantly better than the three baseline methods including naive Tree-RAG.\\ \\
Theoretically, the time complexity of Cuckoo Filter for searching entities is O(1), which is significantly lower than that of naive Tree-RAG. From a spatial point of view, entities are stored in the Cuckoo Filter in the form of fingerprints (12-bit), which greatly saves memory usage. On the other hand, when the load factor of Cuckoo Filter exceeds the preset threshold, the storage capacity is usually increased by double expansion, while the original elements are re-hashed and migrated to the new storage location to complete the automatic expansion. This keeps the loading rate of cuckoo filter high and not too high, thus saving memory while avoiding hash collisions as much as possible.\\ \\
Moreover, we propose two novel designs. The first design introduces a temperature variable, with each entity stored in the Cuckoo Filter maintaining an additional variable called temperature. The variable is used to record the frequency of the entity being accessed. The Cuckoo Filter sorts the entities according to the frequency, and the entities with the highest temperature are placed in the front of the bucket, thus speeding up the retrieval. The second design is to introduce block linked list, where Cuckoo Filter stores the addresses of entities at different locations in the tree. The utilization of the space of block linked list is high, it can support relatively efficient random access, reduce the number of linked list nodes, and perform well in balancing time and space complexity, especially for processing large-scale data. Therefore, we achieve acceleration by storing these addresses in the form of a block linked list. An ablation experiment is performed to demonstrate the effectiveness of the design. \\ \\

\subsection{Related Work}

\label{sec:headings}
\textbf{Bloom Filter} Bloom Filter is an efficient probabilistic data structure for set membership queries~\cite{bloom1970space}. The structure maps elements to bit arrays by using multiple hash functions to quickly determine whether an element exists. On insertion, the locations corresponding to all hash functions are set to 1; in query, if all relevant bits are 1, the element may exist, and if any bit is 0, it must not exist. This design results in false positives but no false negatives for bloom filters. Bloom filters are widely used in scenarios such as databases, caches and distributed systems to quickly filter irrelevant candidates, thus speeding up the retrieval process~\cite{patgiri2019role}. Despite its spatial efficiency, the fixed false positive rate and the limitation of not being able to delete elements make it limited in some occasions.\\ \\
\textbf{Cuckoo Filter} Cuckoo Filter is an efficient data structure for supporting fast element lookup and deletion operations, which is mainly used for collection operations and data stream processing~\cite{fan2014cuckoo}. It is developed based on the idea of Cuckoo Hashing~\cite{pagh2001cuckoo}. Cuckoo Filter outperforms traditional Bloom filters in terms of storage efficiency and query performance, especially in scenarios where frequent insertion and deletion of elements are required. The main advantage of Cuckoo Filter is its support for dynamic updates. The main advantage of Cuckoo Filter is its support for dynamic updates, which enables it to efficiently handle element changes in a collection. Unlike Bloom Filter, Cuckoo Filter can not only query whether an element exists or not, but also support the deletion operation of an element, a feature that is important in many practical applications~\cite{gupta2015cuckoo}. The working principle of Cuckoo Filter is based on multiple hash functions and a bucket structure, which is utilized by storing the elements in fixed-size buckets and using the hash conflicts to achieve fast lookup of elements. This efficient data structure provides new ideas for handling large-scale datasets, which can accelerate the retrieval process and improve response efficiency. \\ \\
\textbf{Retrieval Augmented Generation} Retrieval Augmented Generation(RAG) is a state-of-the-art method that combines information retrieval with large language models, with the aim of addressing the limitations of it in the absence of specific knowledge~\cite{patrick2020retrieval}. The core idea of RAG is to utilize an external knowledge base for retrieval and to incorporate relevant information into the generation process. Specifically, RAG first retrieves multiple relevant documents from the knowledge base on the basis of the input query, and then combines the retrieved knowledge with the input query to form the context and then generate the final answer through the generative model. This approach significantly improves the quality and relevance of the generated text~\cite{karpukhin2020dense} and avoids the limitations of fixed model knowledge~\cite{gupta2024comprehensive}. Compared to fine-tuning, RAG has a greater ability to update knowledge and also reduces the dependence on large-scale data.\\  \\
\textbf{Graph-RAG}  Graph Retrieval-Augmented Generation (Graph-RAG) is an extension of RAG, where the information retrieval process is augmented by leveraging graph-based structures to organize and retrieve information~\cite{patrick2020retrieval}. The key difference between traditional RAG and Graph-RAG is the use of a graph, such as a knowledge graph, to model relationships between entities and concepts, which can improve the relevance and contextuality of the retrieved information~\cite{hu2024graggraphretrievalaugmentedgeneration}. The algorithmic process of Graph-RAG enhances the understanding of relationships and provides more precise information retrieval than traditional RAG~\cite{Darrin2024from}. However, the complexity in graph construction and maintenance will be a trouble, the quality and completeness of the graph can also affect the accuracy of responses generated by the model. Therefore, in comparison to Tree-RAG, graph-RAG still has certain disadvantages.\\ \\
\textbf{Tree-RAG}  Tree-RAG(T-RAG) is an emerging method that combines tree structure and large language models to improve the effectiveness of knowledge retrieval and generation tasks. Compared to traditional RAG, T-RAG further enhances the context retrieved from vector databases by introducing a tree data structure to represent the hierarchy of entities in an organization. The algorithmic process of Tree-RAG consists of the following steps: first, the input query is parsed to identify relevant entities and the retrieval of relevant entities is performed in the constructed forest. Next, the system traverses through the hierarchical structure of the tree to obtain the nodes related to the query entity and its upper and lower multilevel parent-child nodes. Subsequently, the retrieved knowledge is fused with the query to generate the augmented context. Finally, the generative model generates the final answer based on the augmented context. This process effectively combines knowledge retrieval and generation and improves the accuracy and contextual relevance of the generative model~\cite{fatehkia2024t}. However, T-RAG runs inefficiently due to the time-consuming nature of finding all the locations of related entities in a forest with a large amount of data. Our method applies the improved Cuckoo Filter to the retrieval process of Tree-RAG, making it greatly faster.

\section{Data Pre-processing}

\label{sec:others}
It is important to recognize entities and construct hierarchical relationships (e.g., tree diagrams) between entities from datasets. It mainly involves the steps of entity recognition, relationship extraction and filtering.
For existing hierarchical data, binary pairs representing parent-child relationships are directly extracted. For raw textual data, text cleansing is first performed manually to remove irrelevant information.
\subsection{Entities Recognition}
SpaCy is a Python library, and its entity recognition function is based on deep learning models (e.g., CNN and Transformer). It captures information by transforming the text into word vectors and feature vectors. The models are trained on a labeled corpus to recognize named entities in the text, such as names of people and places. We adopt the method in T-RAG by using the spaCy library to recognize and extract entities from a user's query~\cite{fatehkia2024t}. 

\subsection{Relationship Extraction}
\begin{figure}[htbp]
    \centering
    \includegraphics[width=1.0\columnwidth]{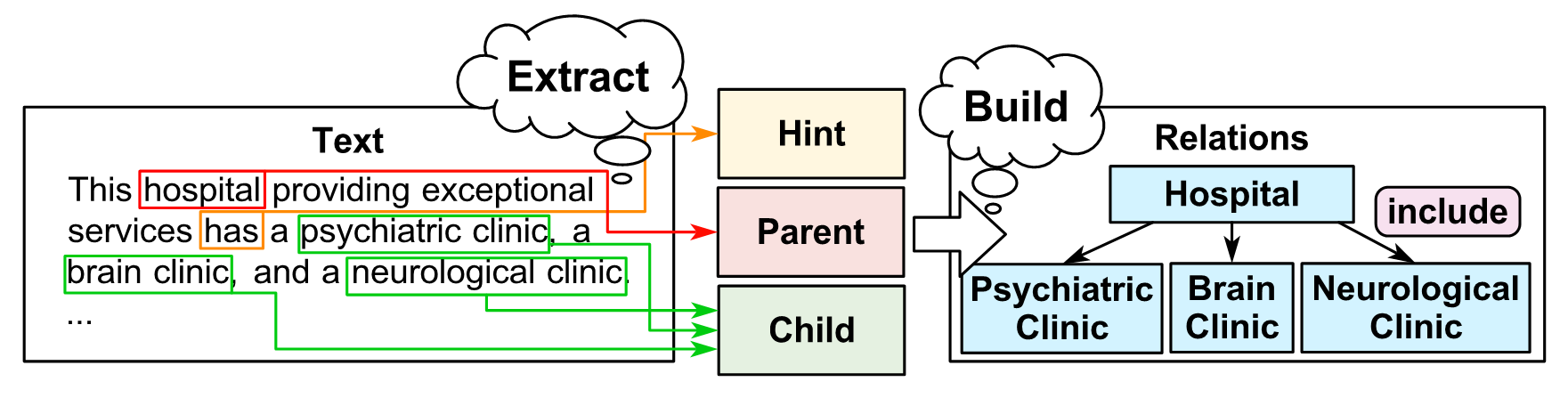}
    \caption{The process of relationship extraction}
    \label{fig:relation}
\end{figure}

Various relationships are identified from the data, including organizational, categorization, temporal, geographic, inclusion, functional, and attribute relationships. The relationships manifest through grammatical structures such as noun phrases, prepositional phrases, relative clauses, and appositive structures ~\cite{vaswani2017attention,devlin2019bert}. We focus on extracting relationships that express dependency, such as "belongs to," "contains," and "is dependent on." The process is presented in Figure \ref{fig:relation}.

We use several dependency parsing models(gpt-4 and open-source NLP libraries) to analyze the grammatical structure of the data. This helps identify relationships between words, such as subject-verb-object or modifier relationships.

We define rules to identify hierarchical relationships. If a word modifies another noun, it can be interpreted as a child-parent relationship; If there are conjunctions (e.g., "and", "or"), handle them to group entities under the same parent. As a result, there is a list of tuples representing the hierarchical structure.

\subsection{Relationship Filtering}
After extracting relations, certain relationships are filtered out to ensure maintain the tree structure:\\ 
\begin{figure}[!h]
    \centering
    \includegraphics[width=1.0\columnwidth]{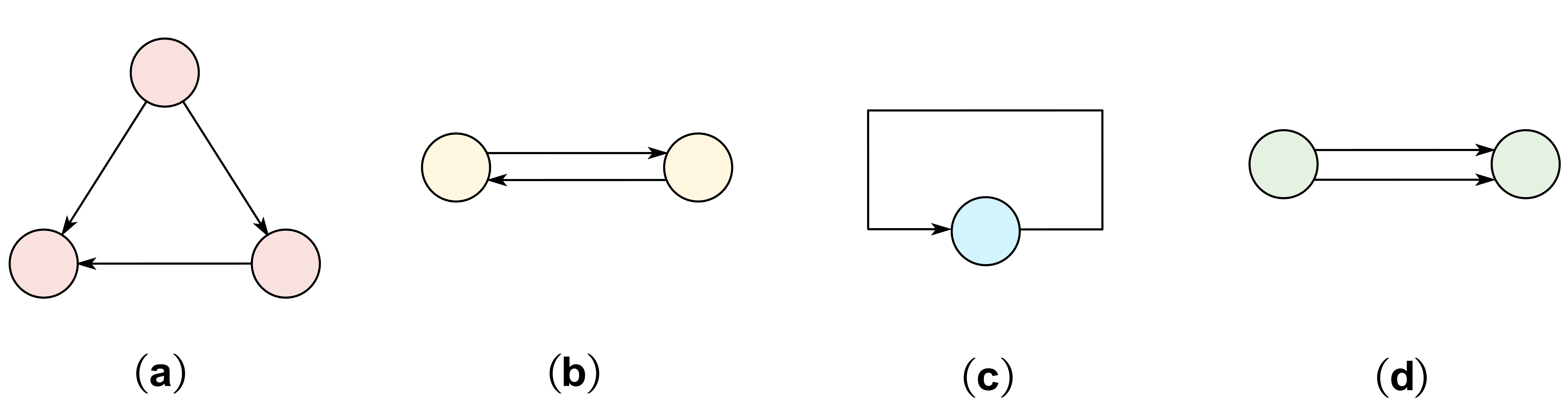}
    \caption{Error relation examples.}
    \label{fig:my_label}
\end{figure}
\begin{itemize}
\item Transitive Relations: If transitive relations are detected (e.g., "A belongs to B", "B belongs to C" and "A belongs to C"), remove distant relations.
\item Cycle Relations: If cycles are detected (e.g., "A belongs to B" and "B belongs to A"), only the closest relationship is retained.
\item Self-Pointing Edges: Any relation where a node points to itself is removed.
\item Duplicate Edges: Multiple edges between the same nodes are pruned, leaving only one edge. 
\end{itemize}

\section{Methodology}
In this section, we propose a novel design of Cuckoo Filter that combines the advantages of traditional Cuckoo Hashing and applies it to Tree-RAG by introducing additional designs that greatly improve the speed of knowledge retrieval in Tree-RAG.

\subsection{Storage Mode} 
\begin{figure*}[htbp]
    \centering
    \includegraphics[width=\textwidth]{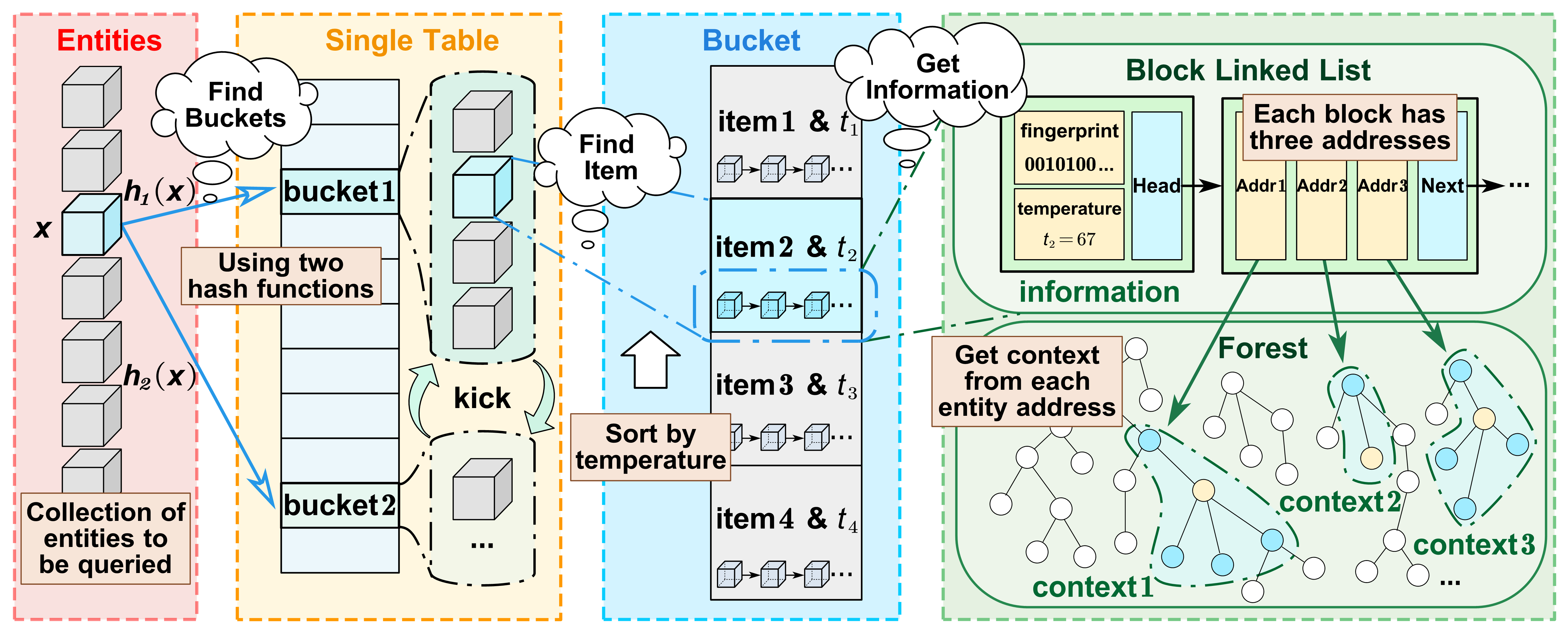}
    \caption{The workflow of CFT-RAG when query contains entity x. The entity with high temperature will be placed ahead of which with low temperature in the bucket. All the addresses in different trees of the entity are linked by the block 
 linked list.}
    
    \label{fig:my_label2}
\end{figure*}

In addition to entity trees, we set up an additional Cuckoo Filter to store some entities to improve retrieval efficiency. Based on the naive Cuckoo Filter, we introduce the block linked list for optimization, which can greatly reduce memory fragmentation. We first find out all locations of each entity in the forest and then store these addresses in a block linked list. \\ \\
To further optimize the retrieval performance, we propose an adaptive sorting strategy to reorder the entities in each bucket in the Cuckoo Filter based on the temperature variable which is stored at the head of the block list. The temperature variable records how often each entity is accessed, and entities with high-frequency access are prioritized to be placed at the front of the bucket. Since the Cuckoo Filter looks up the elements in the buckets linearly, this reordering mechanism can significantly optimize the query process, which can further improve the response speed of the model. In summary, in each entry of the bucket, an entity's fingerprint, its temperature, and head pointer of its block linked list are stored. The storage mode is included in Figure \ref{fig:my_label2}.

\subsection{Entity Insertion Strategy}
The fingerprint is a shorter hash representation of an entity x, which is usually represented in fixed-length bits. Thus, we introduce fingerprints to save memory. After the tree structure is constructed, the fingerprint is first computed for each entity before inserting. Furthermore, the insertion and eviction strategy is consistent with the traditional Cuckoo Filter, where the locations of the fingerprint are determined by two hash functions~\cite{fan2014cuckoo}.
When inserting an entity by applying the Cuckoo Filter, it first tries to store its fingerprint in the empty position $i_{1}$ or $i_2$, which can be calculated as:

\begin{equation}
i_1 = h(x),
i_2 = i_1 \oplus h(f(x))
\end{equation}

where $h(x)$ is the hash value of entity x and $h(f(x))$ is the hash value of fingerprint $f(x)$. The two locations $i_1$ and $i_2$ are candidate storage locations to increase the flexibility of the lookup and insertion operations. 
\begin{algorithm}[H]
\SetAlgoNoLine  
        \caption{Entity Insertion Algorithm}
        $f(x)$ = fingerprint(x);
                
        \If{either bucket[$i_1$] or bucket[$i_2$] has an empty entry}{
            add $f(x)$ and the initialized temperature t to that bucket; \\
            add the addresses of the entity as block linked list to that bucket;
            
            \Return{True} }
        randomly select $i_1$ or $i_2$ as i;\\
        \For{$k=0;k < MaxNumKicks;k ++ $}
    {
        select an entry m from bucket[i] randomly;  \\
        interchange $f(x)$ and the fingerprint stored in entry m;  \\
 i = i $\oplus$ hash($f(x)$); 
 \\\If{bucket[i] has an empty entry}{
 add $f(x)$ and the initialized temperature t to bucket[i];  \\
 add the addresses of the entity as block linked list to bucket[i]; \\
 \Return{True}}
    }  
    \Return{False} 
\end{algorithm}

Then we check if there are empty slots at locations $i_1$ and $i_2$. If an empty slot exists in either position, the information of entity x including its fingerprint $f(x)$, its temperature t and its addresses are inserted into that empty slot and the insertion process is completed. If both locations are occupied, the Cuckoo Filter activates the eviction mechanism and tries to free up space for inserting a new entity by evicting the existing entity. In practice, the Cuckoo Filter sets the maximum number of evictions to avoid infinite recursive eviction operations; after reaching the maximum number of evictions, it means that the insertion operation fails.

\subsection{Entity Eviction Strategy}
In the eviction mechanism, the Cuckoo Filter randomly selects a location from $i_1$ or $i_2$ to evict the fingerprints therein and then calculate the alternative position j of the fingerprint. Next, it inserts the fingerprint into the empty slot of j. If it is already occupied, the eviction operation is repeated until an empty location is found or the maximum number of iterations is reached.

\begin{algorithm}[H]
 \SetAlgoNoLine  
    \caption{Entity Eviction Algorithm}
         $f(x)$ =fingerprint(x);\\
 $i_1$ = hash(x);\\
 $i_2$ = $i_1$ $\oplus$ hash($f(x)$);\\
 \If{ bucket[$i_1$] or bucket[$i_2$] has $f(x)$}{
 remove $f(x)$ together with its information from this bucket;\\
 \Return{True}}
 \Return{False}

\end{algorithm}

Through the eviction mechanism, Cuckoo Filter is able to continue inserting new entities under high load conditions, avoiding the high misclassification rate caused by direct rejection of insertion.

\begin{algorithm}[H]
\caption{Context Generation Algorithm}
\SetAlgoNoLine
\SetAlgoNlRelativeSize{0} 

\SetAlgoNlRelativeSize{0}

\KwIn{$x$: Input entity}
\KwOut{$context$: Context generated for entity $x$}

$f(x) \gets \text{fingerprint}(x);$\;\\
\If{bucket[$i_1$] or bucket[$i_2$] contains $f(x)$}{
    $temperature \gets temperature + \text{1};$\;\\
    \Return{head}\;
}

$currentBlock \gets \text{head} \rightarrow \text{next};$\;

\While{$currentBlock \neq \text{NULL}$}{
    
    \ForEach{location in currentBlock}{
       Let $loc$ be the current location of entity $x$ in the block; \\
       Find the set of hierarchical relationship nodes at location $loc$ in the tree; \\
        $H_{up} \gets \{h_1, h_2, \dots, h_n\};$\; \\ 
        $H_{down} \gets \{h'_1, h'_2, \dots, h'_n\};$\;  \\
       Record the first $n$ upward and downward hierarchical relationship nodes; \\
        \For {($i=1$ ; $i$ < $n$ ; $i$++)}{$\quad \text{Store $(h_i, h'_i)$ in context;}$;}
        
    }
    $currentBlock \gets currentBlock \rightarrow \text{next};$
}
 \For{($i=1$ ; $i < n$ ; $i++$)}{$context \gets context \cup (h_i, h'_i)$;}

\end{algorithm}

\subsection{Lookup Process and Context Generation}
When an entity needs to be looked up, its location is calculated in the same way as $i_1$ and $i_2$ above. Therefore, we only need to check these two locations for the presence of the entity.
If the fingerprint of the target entity is found, the temperature of the entity is added by one and a pointer to the head of the corresponding block linked list of that entity is returned. From this pointer, the location of the entity node in different trees including multi-level parent nodes, child nodes, etc. can be accessed through the address stored in the block list. If no matching fingerprint is found, the null pointer is returned. For the queried entity and its parent and child nodes in different trees, we form a context between the entity and its relevant nodes based on the set template. For instance, the upward hierarchical relationship of entity A are: B, C and D. Finally, we fuse this information with the query to generate the augmented context. After that, the augmented context combined with system prompt and query is regarded as the prompt. The lookup and context generation process is stated in Figure \ref{fig:my_label2}.

\section{Experiments}
\label{sec:experiments}

In this section, we describe our experimental setup, baseline methods, and Cuckoo Filter T-RAG, and comprehensively evaluate the results. Our experiments aim to assess the efficiency of Cuckoo Filter T-RAG approaches under diverse experimental conditions, particularly in terms of retrieval speed and computational overhead.

\subsection{Baseline}
To benchmark the effectiveness of the proposed Cuckoo Filter T-RAG, we compare it against several baseline models. Our baselines include the Naive T-RAG without filtering mechanisms, T-RAG with Bloom Filter, and Improved Bloom Filter T-RAG. These baselines allow us to quantify the improvements introduced by Cuckoo Filter T-RAG.

\textbf{Naive T-RAG}
This basic implementation of T-RAG does not include any filtering optimizations. The method constructs an entity tree using entities extracted from the dataset and employs a Breadth-First Search (BFS) algorithm for entity lookup. Although this approach has high time complexity and prolonged search time, it provides a straightforward baseline for evaluating the benefits of incorporating filtering mechanisms.

\textbf{Bloom Filter T-RAG}
In this model, we incorporate a Bloom Filter at each node in the entity tree. The Bloom Filter of each node indicates whether an entity exists in the node or its descendants. During retrieval, if a Bloom Filter suggests that an entity is absent, the search path is pruned, thereby reducing the search time significantly. This model serves to evaluate the potential time savings achievable through probabilistic filtering.

\textbf{Improved Bloom Filter T-RAG}
Building upon the Bloom Filter T-RAG, we optimize Bloom Filter usage by skipping Bloom Filter checks at nodes just above the leaf level. This change reduces unnecessary filter operations, lowering the time complexity further while maintaining high retrieval accuracy. The efficiency gains from this improvement are compared against the standard Bloom Filter T-RAG to showcase its effectiveness.

\subsection{Cuckoo Filter T-RAG}
Using Cuckoo Filter has more prominent advantages including time efficiency than Bloom Filter. Cuckoo Filter supports entity deletion operation, which is suitable for ongoing data update, and it has a lower false positive rate and is more space efficient.

The Cuckoo Filter T-RAG method stores the individual nodes of the entity in the forest in each bucket of the Cuckoo Filter, i.e. it merges the Cuckoo Hash with the Cuckoo Filter. After the entity tree is generated, the nodes with the same entity details in each tree are concatenated into a block list, where the pointer to the head of the list corresponds to the fingerprint, and stored together in buckets.

An improved Cuckoo Filter T-RAG is to maintain access popularity of each entity, called temperature, at the head node of each block list, and raise the level of temperature corresponding to the hit entity during retrieval. For each bucket, if there is a bucket that has not been searched, i.e. if it is free, the fingerprints and block list header pointers in the bucket can be sorted according to temperature, and the fingerprints with higher access popularity are placed at the front of the bucket, which can take advantage of the locality of the entities contained in the user questions to improve the running speed of the algorithm.

\subsection{Datasets and Entity Forest}
Our experiments use two datasets: the English dataset of the UNHCR organizational chart mentioned in T-RAG and a real-world Chinese dataset of hospital histories. The UNHCR dataset is pre-segmented into entities, allowing us to focus on entity extraction from the hospital histories dataset. We leverage
dependency parsing models
to extract entities and relationships among them and construct the entity forest based on these extracted entities and relationships. The resulting entity forest is structured to allow efficient retrieval and provides a practical evaluation scenario for our approach.

\subsection{Setup}
We implement the core RAG architecture in Python, while key data structures, including the Bloom and Cuckoo Filters, are optimized in C++ for performance. All experiments were conducted on a system equipped with an Nvidia H100 GPU, 22 CPU cores, and 220 GiB of memory. Each algorithm was repeated 100 times to account for variability and ensure reliable results, with averages calculated across runs to mitigate the influence of outliers. We apply the langsmith framework to evaluate the accuracy of answers, where the OpenAI scoring model used by langsmith was replaced with doubao.  

\subsection{Results and Evaluations}

\subsubsection{Comparison Experiment}

Table~\ref{tab:table} summarizes the performance metrics for each model when the number of trees is different. In the experiments comparing the retrieval speeds of the different methods, only a small number of entities are included in the user query (the number of entities in the query is set to 5, 10, and 20). By integrating the Bloom Filter, retrieval time is partially reduced, as the filter prunes non-relevant paths in the entity tree. The improved Bloom Filter T-RAG achieves an additional speedup, attributed to the selective omission of Bloom Filter checks on near-leaf nodes, which minimizes overhead without sacrificing accuracy. From Table~\ref{tab:table}, we can observe that the optimization range increases with the number of trees in the database. For example, when the tree number is 600, our method is 138 times faster than naive Tree-RAG while maintaining the same accuracy.

 \begin{table}[H]
   \centering
   \begin{threeparttable}
     \caption{Retrieval time of each algorithm}
     \begin{tabular}{clcc}
       \toprule
       Tree Number         & Algorithm                      & Time(s)       & Acc(\%)   \\
       \midrule
       \multirow{4}{*}{50} & Naive T-RAG                    & 0.320925      & 66.30            \\
                           & BF\phantom{ive }T-RAG          & 0.250479      & 66.70            \\
                           & BF2\phantom{v.} T-RAG          & 0.188924      & 66.58            \\
                           & CF\phantom{ive} T-RAG          & \textbf{0.011772}      & 66.45            \\
       \midrule
       \multirow{4}{*}{300} & Naive T-RAG                    & 1.767487     & 66.75             \\
                            & BF\phantom{ive }T-RAG          & 1.360986     & 66.30            \\
                            & BF2\phantom{v.} T-RAG          & 1.097023     & 65.70            \\
                            & CF\phantom{ive} T-RAG          & \textbf{0.017502}     & 66.28            \\
       \midrule
       \multirow{4}{*}{600}  & Naive T-RAG                    & 3.301436     & 66.45            \\
                             & BF\phantom{ive }T-RAG          & 2.638721     & 66.60            \\
                             & BF2\phantom{v.} T-RAG          & 2.171958     & 65.80            \\
                             & CF\phantom{ive} T-RAG          & \textbf{0.023776}     & 66.50            \\
       \bottomrule
     \end{tabular}
     \begin{tablenotes}[flushleft]
       \scriptsize
       \item BF T-RAG represents Bloom Filter T-RAG.
       \item BF2 T-RAG represents Improved Bloom Filter T-RAG.
       \item CF T-RAG represents Cuckoo Filter T-RAG.
       \item Acc represents the model's accuracy of answer.
     \end{tablenotes}
     \label{tab:table}
   \end{threeparttable}
 \end{table}

 \begin{table}[H]
  \centering
  \begin{threeparttable}
    \caption{Retrieval time of each algorithm when different number of entities included in the query (600 trees)}
    \begin{tabular}{clcc}
      \toprule
      Entity Number         & Algorithm                      & Time(s)       & Acc(\%)   \\
      \midrule
      \multirow{4}{*}{5} & Naive T-RAG                    & 12.615021      & 66.45            \\
                          & BF\phantom{ive }T-RAG          & 6.568509      & 66.60            \\
                          & BF2\phantom{v.} T-RAG          & 4.704859      & 66.80            \\
                          & CF\phantom{ive} T-RAG          & \textbf{0.074715}      & 66.50            \\
      \midrule
      \multirow{4}{*}{10} & Naive T-RAG                    & 13.673550     & 66.45             \\
                           & BF\phantom{ive }T-RAG          & 8.700830     & 66.60            \\
                           & BF2\phantom{v.} T-RAG          & 5.256318     & 65.80            \\
                           & CF\phantom{ive} T-RAG          & \textbf{0.090924}     & 66.50            \\
      \midrule
      \multirow{4}{*}{20}  & Naive T-RAG                    & 38.479844     & 66.45            \\
                            & BF\phantom{ive }T-RAG          & 11.146254     & 66.60            \\
                            & BF2\phantom{v.} T-RAG          & 10.461809     & 65.80            \\
                            & CF\phantom{ive} T-RAG          & \textbf{0.087526}    & 66.50            \\
      \bottomrule
    \end{tabular}
    \begin{tablenotes}[flushleft]
      \scriptsize
      \item BF T-RAG represents Bloom Filter T-RAG.
      \item BF2 T-RAG represents Improved Bloom Filter T-RAG.
      \item CF T-RAG represents Cuckoo Filter T-RAG.
      \item Acc represents the model's accuracy of answer.
    \end{tablenotes}
    \label{tab:table2}
  \end{threeparttable}
\end{table}

Table~\ref{tab:table2} summarizes the performance metrics for each model when the number of entities included in the query is different. In addition, we can see that a user's query often contains more than one entity in real life, especially in a long-text query, which often contains a large number of entities. Therefore, the speed of processing long text queries containing multiple entities is also significant. From Table~\ref{tab:table2}, we can observe that the retrieval time of the three baseline methods increases significantly as the number of entities in the query increases while the retrieval time of our method remains stable.

\begin{figure}[h!]
    \centering
    \includegraphics[width=1.0\columnwidth]{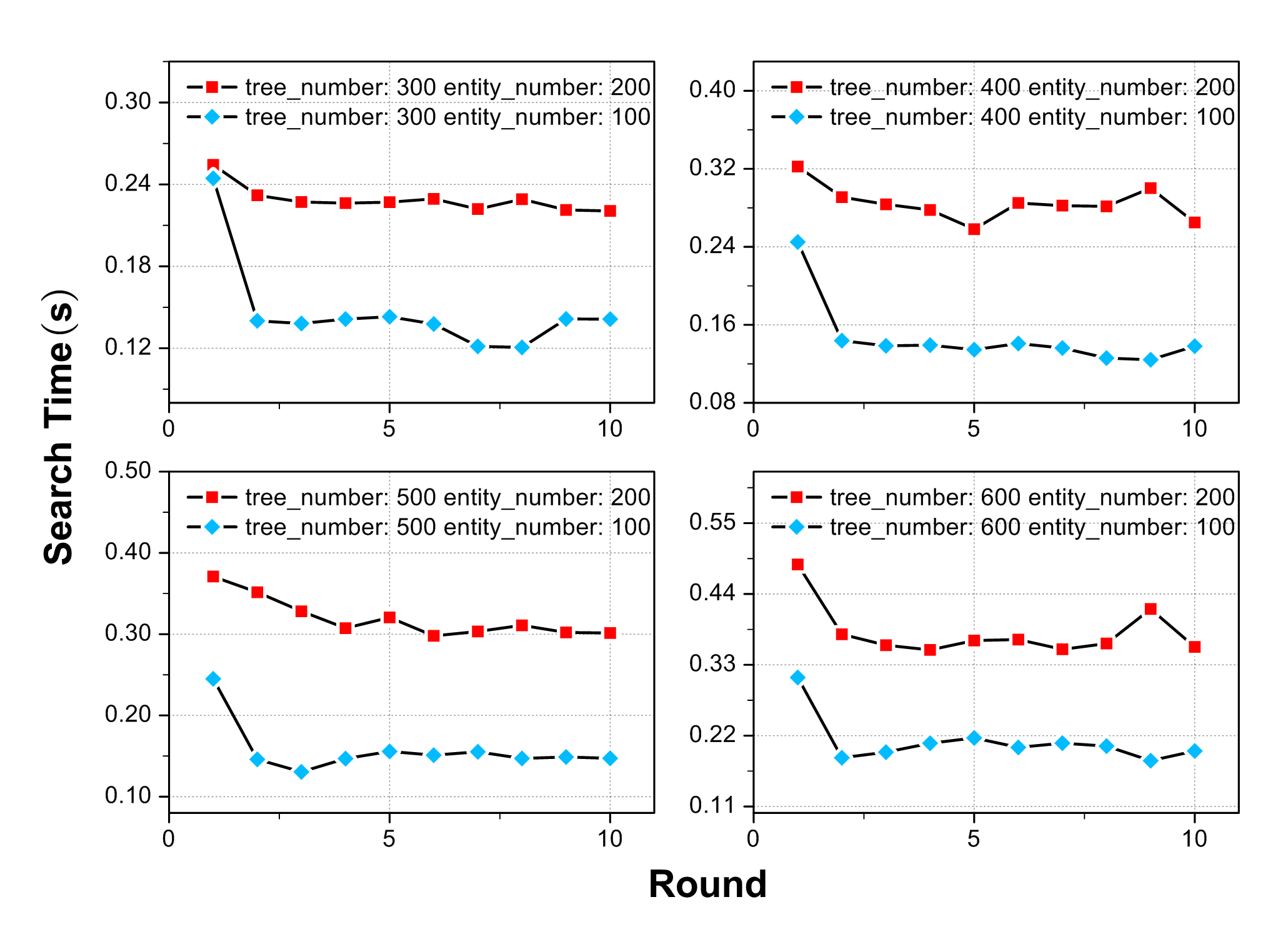}
    \caption{We record the search time per round of query with different number of trees and entities. Each round represents a search in the entities forest, and entities are inserted into the improved Cuckoo Filter before the first search is performed.}
    
    \label{fig:sort_label}
\end{figure}

Moreover, the error rate of our method in the process of searching entities is very low. After building the trees based on the hospital dataset, the Cuckoo Filter includes 1024 buckets, each of which can hold up to 4 fingerprints and block linked list head pointers. The Cuckoo Filter's own memory expansion strategy is to increase the number of buckets by a power of two. In the experimental dataset, there are 3,148 entities that can be extracted, and the space load factor for the Cuckoo Filter is 0.7686. Because the space load factor is not too high and searching errors are mainly caused by hash collisions, the error rate is almost 0, showing that the number of entities causing the lookup error is 0 to 1 out of 1024 buckets for 3148 entities.

\subsubsection{Ablation Experiment}
Sorting the fingerprints and head pointers of the block linked lists by temperature can optimize the retrieval time without occupying any extra space, which is eminently useful when the query given by the user contains a large number of entities.

We design experiments to measure the effect of having or not having the sorting design on the result. In figure \ref{fig:sort_label}, we can observe that the retrieval time after the first round is significantly shorter than that of the first round. This is because the temperatures are updated according to the access frequency in each round and after each query, the Cuckoo Filter sorts the entities according to the entities' temperatures. This sorting design allows the 'hot' entities to be found more quickly in subsequent queries.

\section{Conclusion}
In this paper, we have introduced an efficient acceleration method for the Tree-RAG framework by integrating the improved Cuckoo Filter into the knowledge retrieval process. Tree-RAG, which combines hierarchical tree structures for knowledge representation with generative models, holds great promise for improving the quality and contextual relevance of generated responses. However, its performance is hindered by the computational inefficiencies of retrieving and organizing large-scale knowledge within complex tree structures.\\ \\
By leveraging the Cuckoo Filter, which supports fast membership queries and dynamic updates, we have significantly enhanced the speed and efficiency of the retrieval process in Tree-RAG. Our experimental results show that the Cuckoo Filter improves retrieval times without sacrificing the quality of generated responses, making the system more scalable for real-world applications. This acceleration is particularly valuable in scenarios where real-time knowledge updates and rapid information retrieval are critical, such as in large-scale question answering, decision support systems, and conversational agents.\\ \\
Future work could explore further optimizations, such as adapting the method for different knowledge structures or extending it to more complex multimodal tasks. Overall, this research demonstrates the potential of efficient data structures to enhance the performance of large language models in retrieval-intensive applications.



\bibliography{example_paper}
\bibliographystyle{icml2025}

\newpage
\appendix
\onecolumn
\section{Appendix}


\end{document}